\documentclass[11pt,a4paper]{article}
\usepackage[hyperref]{emnlp2018}
\usepackage{times}
\usepackage{latexsym}

\usepackage{url}
\usepackage{relsize}
\usepackage{tabularx}
\usepackage{graphicx}
\usepackage{float}
\usepackage{xspace}
\usepackage{footmisc}
\usepackage{lscape}
\usepackage{makecell}

\aclfinalcopy 

\newcommand{\App}{SetExpander\xspace}

\title{Term Set Expansion based NLP Architect by Intel AI Lab}
\author{Jonathan Mamou, Oren Pereg, Moshe Wasserblat,\\
    \textbf{Alon Eirew, Yael Green, Shira Guskin, Peter Izsak, Daniel Korat} \\
  Intel AI Lab, Israel \\
  {\tt firstname.lastname@intel.com} \\
}

\date{}

\begin{document}

\maketitle

\begin{abstract}
We present \App, the term set expansion system based NLP Architect by Intel AI Lab. \App is a corpus-based system for expanding a seed set of terms into a more complete set of terms that belong to the same semantic class. 
It implements an iterative end-to-end workflow and enables users to easily select a seed set of terms, expand it, view the expanded set, validate it, re-expand the validated set and store it, thus simplifying the extraction of domain-specific fine-grained semantic classes. \App has been used successfully in real-life use cases including integration into an automated recruitment system and an issues and defects resolution system.\footnote{A video demo of \App is available at \url{https://drive.google.com/open?id=1e545bB87Autsch36DjnJHmq3HWfSd1Rv} (some images were blurred for privacy reasons).}
\end{abstract}

\section{Introduction}
\label{sec:introduction}
Term set expansion is the task of expanding a given partial set of terms into a more complete set of terms that belong to the same semantic class.  For example, given a seed of personal assistant application terms like `Siri' and `Cortana', the expanded set is expected to include additional terms such as `Amazon Echo' and `Google Now'.
Many NLP-based information extraction applications, such as relation extraction or document matching, require the extraction of terms belonging to fine-grained semantic classes as a basic building block. 
A practical approach to extracting such terms 
is to apply a term set expansion system. The input seed set for such systems may contain as few as 2 to 10 terms, which is practical to obtain.
\App uses a corpus-based approach based on the {\it distributional similarity hypothesis}~\cite{harris1954distributional}, stating that semantically similar words appear in similar contexts. Linear bag-of-words context is widely used to compute semantic similarity.
However, it typically captures more {\it topical} and less {\it functional} similarity, while for the purpose of set expansion, we need to capture more functional and less topical similarity.\footnote{We use the terminology introduced by~\cite{turney2012domain}: the {\it topic} of a term is characterized by the nouns that occur in its neighborhood while the {\it function} of a term is characterized by the syntactic context that relates it to the verbs that occur in its neighborhood.} For example, given a seed term like the programming language 'Python', we would like the expanded set to include other programming languages with similar characteristics, but we would not like it to include terms like `bytecode' or `high-level programming language' despite these terms being semantically related to `Python' in linear bag-of-words contexts.
Moreover, for the purpose of set expansion, a seed set contains more than one term and the terms of the expanded set are expected to be as functionally similar to {\it all} the terms of the seed set as possible.
For example, `orange' is functionally similar to `red' (color) and to `apple' (fruit), but if the seed set contains both `orange' and `yellow' then only `red' should be part of the expanded set. However, we do not want to capture only the term sense; we also wish to capture the granularity within a category. For example, `orange' is functionally similar to both `apple' and `lemon'; however, if the seed set contains `orange' and `banana' (fruits), the expanded set is expected to contain both `apple' and `lemon'; but if the seed set is `orange' and `grapefruit' (citrus fruits), then the expanded set is expected to contain `lemon' but not `apple'.

While term set expansion has received attention from both industry and academia, there are only a handful of available implementations. Relative to prior work, the contribution of this paper is twofold. First, it presents an iterative end-to-end workflow that enables users to select an input corpus, train multiple embedding models and combine them; after which the user can easily select a seed set of terms, expand it, view the expanded set, validate it, iteratively re-expand the validated set and store it.
Second, it describes the \App application that provides these abilities. \App is based on a novel corpus-based set expansion algorithm. This algorithm combines multi-context term embeddings using a neural classifier in order to capture different aspects of semantic similarity and to make the system more robust across different semantic classes and different domains. The algorithm is briefly described in Section~\ref{sec:algo}.
Our system has been used successfully in several real-life use cases. One of them is an automated recruitment system that matches job descriptions with job-applicant resumes.
Another use case involves enhancing a software development process by detecting and reducing the amount of duplicate defects in a validation system. Section~\ref{sec:usecases} includes a detailed description of both use cases.
The system is distributed as open source software under the Apache license as part of NLP Architect by Intel AI Lab.\footnote{\url{http://nlp_architect.nervanasys.com/term_set_expansion.html}}

\section{Related Work}

State-of-the-art set expansion techniques return the $k$ nearest neighbors around the seed terms as the expanded set, where terms are represented by their co-occurrence or embedding vectors in a training corpus. Vectors are constructed according to different context types, such as linear bag-of-words context~\cite{pantel2009web,shi2010corpus,rong2016egoset,zaheer2017deep,gyllensten2018distributional}, explicit lists~\cite{roark1998noun,sarmento2007more,he2011seisa}, coordinational patterns~\cite{sarmento2007more} and unary patterns~\cite{rong2016egoset,shen2017setexpan}.
\App looks at additional context types that can capture functional semantic similarities and combines context type embeddings using a neural classifier. 

Google Sets, now discontinued, was one of the earliest web applications for term set expansion. It used methods like latent semantic indexing to pre-compute lists of similar words from the web. 
Word Grab Bag\footnote{\url{www.wordgrabbag.com}} is another web application based on a method that builds lists dynamically using word2vec embeddings based on linear bag-of-word contexts, but their algorithm is not publicly described. 
Later, \citet{wang2007language} proposed the SEAL (Set Expander for Any Language) system which automatically finds semi-structured web pages that contain `lists of' items, and then aggregates these lists so that the most promising items are ranked higher.
In our paper, we describe an iterative end-to-end system, including model training and using additional context types.

\citet{pantel2009web} propose a highly scalable algorithm, implemented in the MapReduce framework, for computing semantic similarity, where terms are represented by large and sparse co-occurrence vectors. \App ensures scalability by representing terms with small and dense embeddings vectors. 

The current paper extends \cite{mamou2018setexpander} paper.

\section{Algorithm Overview}
\label{sec:algo}

\subsection{Term Extraction and Representation}
\label{sec:TermRepresentation}

Our approach is based on representing any term of an (unlabeled) training corpus by its word embeddings in order to estimate the similarity between seed terms and candidate expansion terms. 

Noun phrases provide good approximation for candidate terms and are extracted in our system using a noun phrase chunker.\footnote{\url{http://nlp_architect.nervanasys.com/chunker.html}} Term variations, such as aliases, acronyms and synonyms, which refer to the same entity, are grouped together.\footnote{For that, we use a heuristic algorithm based on text normalization, abbreviation web resources, edit distance and word2vec similarity. For example, {\it New York, New-York, NY, NYC} and {\it New York City} are grouped.} 
Next, we use term groups as input units for embedding training; this enables obtaining more contextual information compared to using individual terms, thus enhancing embedding model robustness. In the remainder of this paper, by language abuse, {\it term} will be used instead of term group. 

While word2vec originally uses a linear bag-of-words context around the {\it focus term} to learn the term embeddings, the literature describes other possible context types. For each focus term, we extract {\it context units} of different types, as follows (see examples in Table~\ref{table:context:example}). 

\begin{table*}
\begin{tabularx}{\textwidth}{|l|X|l|X|}
    \hline
    
    \bf{Context} &  \bf{Example sentence} & \bf{Focus} & \bf{Context units} \\ 
    \bf{type} &   & \bf{term} &  \\ \hline
    Linear & {\it Siri uses voice queries and a  } & {\it Siri} & {\it uses, voice queries, natural }\\ 
    $win=5$ &{\it natural language user interface.} & &  {\it language user interface} \\ \hline
    List & {\it Experience in Image processing,} & {\it Image} & {\it Signal processing, Computer Vision} \\ 
     & {\it Signal processing, Computer Vision.} & {\it processing} &  \\ \hline
    Dep & {\it Turing studied as an undergraduate ... at King's College, Cambridge.} & {\it studied} & {\it (Turing/nsubj), (undergraduate\-/\-prep\_as), (King's College/prep\_at)} \\ \hline
    SP & {\it Apple and Orange juice drink ... } & {\it Apple} & {\it Orange} \\ \hline
    UP & {\it In the U.S. state of Alaska ...} & {\it Alaska} & {\it U.S. state of \_\_} \\ \hline
\end{tabularx}
\caption{Examples of extracted contexts per context type.}
\label{table:context:example}
\end{table*}

\paragraph{Linear Bag-of-Words Context.} This context type is defined by neighboring context units within a fixed length window of context units, denoted by $win$, around the focus term. Both terms and other words can be context units. One of its implementations is word2vec~\cite{mikolov2013distributed}, widely used for NLP tasks including set expansion.

\paragraph{Explicit Lists.} Context units consist of terms co-occurring with the focus term in textual lists such as comma separated lists and bullet lists~\cite{roark1998noun}. 

\paragraph{Syntactic Dependency Context (Dep).} This context type is defined by the syntactic dependency relations in which the focus term participates~\cite{levy2014dependency}. Context units consist of terms or other words, along with the type and the direction of the dependency relation.
This context type has not been used for set expansion in prior work. However, \citet{levy2014dependency} showed that this context yields more functional similarities of a co-hyponym nature than is yielded by linear bag-of-words context, which suggests its relevance for set expansion.

\paragraph{Symmetric Patterns (SP).} Context units consist of terms co-occurring with the focus term in symmetric patterns~\cite{davidov2006efficient}. For example, the symmetric pattern `X rather than Y' captures a certain semantic relatedness between the terms X and Y. This context type generalizes coordinational  patterns (`X and Y', `X or Y'), which have been used for set expansion.

\paragraph{Unary Patterns (UP).} This context type is defined by the unary patterns in which the focus term occurs~\cite{rong2016egoset}. Context units consist of $n$-grams of terms and other words, in which the focus term occurs; `\_\_' denotes the placeholder of the focus term in Table~\ref{table:context:example}.\footnote{Following~\citet{rong2016egoset}, we extract six $n$-grams per focus term. Given a sentence fragment $c_{-3} \; c_{-2} \; c_{-1} \; t \; c_1 \; c_2 \; c_3$ where $t$ is the focus term and $c_i$ are the context units, the following $n$-grams are extracted: 
$(c_{-3} \; c_{-2} \; c_{-1} \; t \; c_1)$, 
$(c_{-2} \; c_{-1} \; t \; c_1 \; c_2)$,
$(c_{-2} \; c_{-1} \; t \; c_1)$,
$(c_{-1} \; t \; c_1 \; c_2 \; c_3)$, 
$(c_{-1} \; t \; c_1 \; c_2)$,
$(c_{-1} \; t \; c_1)$.} 

\paragraph{} 
We found that indeed in different domains and for different semantic classes, better similarities are found using different context types. The different contexts thus complement each other by capturing different types of semantic relations.
For example, explicit list contexts worked well for the automated recruitment system use case, while unary patterns contexts worked well for the issues and defects resolution use case (discussed in Section~\ref{sec:usecases}). 
Moreover, explicit lists, syntactic dependency, symmetric patterns and unary patterns context types tend to capture functional rather than topical semantic similarities. 
We train a separate term embedding model for each of the five context types and thus, for each term, we obtain five different representations. 

Terms are represented by their linear bag-of-words window context embeddings using the word2vec toolkit~\footnote{\url{http://code.google.com/archive/p/word2vec}}
 and by arbitrary context embeddings using the generic word2vecf toolkit.\footnote{\url{http://bitbucket.org/yoavgo/word2vecf}}. 
For each focus term in the corpus, $<$focus term, context unit$>$ pairs are extracted from the corpus and are then fed to the embeddings training algorithm.
Concerning linear bag-of-words context type, some hyperparameters of the term embeddings training can be tuned to optimize the set expansion task; in particular, a smaller window size seems to capture functional rather than topical semantic similarities~\cite{levy2014dependency}.

\subsection{Multi-Context Term Similarity}
\label{sec:MultiContextSimilarity}
To make set expansion more robust, we aim to combine multi-context embeddings. 
Following~\cite{berant2012learning}, who train a Support Vector Machine (SVM) to combine different similarity score features, we train a Multilayer Perceptron (MLP) classifier that predicts whether a candidate term should be part of the expanded set based on ten similarity scores (considered as input features) obtained by the five different context types and two different similarity-scoring methods. The two similarity scores are estimated by the cosine similarity between the centroid of the seed terms and each candidate term, and by the average pairwise cosine similarity between each seed term and each candidate term; both methods ensure that the candidate term is similar to all the seed terms. MLP is trained on a labeled training set of seed terms and candidate terms.

\subsection{Implementation and Evaluation}

NLP Architect by Intel AI Lab~\footnote{\url{https://github.com/NervanaSystems/nlp-architect}} has been used for noun phrase chunking, dependency parsing and term embeddings model training.
The performance of the algorithm was first evaluated by the Mean Average Precision at different top $n$ values (MAP@$n$). MAP@10, MAP@20 and MAP@50 on an English Wikipedia based dataset~\footnote{Dataset is described at \url{http://nlp_architect.nervanasys.com/term_set_expansion.html}.} are respectively 0.83, 0.74 and 0.63. These figures indicate the quite useful performance of the algorithm, which was further assessed by the use cases described in Section~\ref{sec:usecases}.

\section{System Workflow and Application}
\label{sec:workflow}

\begin{figure*}
\includegraphics[width=0.7\textwidth]{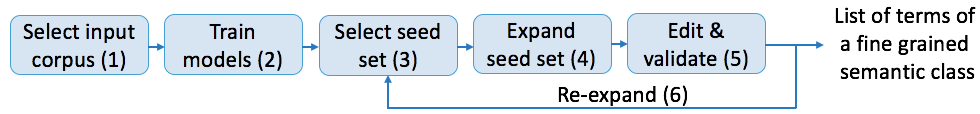}
\centering
\caption{\App end-to-end workflow.}
\label{flow_fig}
\end{figure*}

\begin{figure*}

\includegraphics[width=0.7\textwidth]{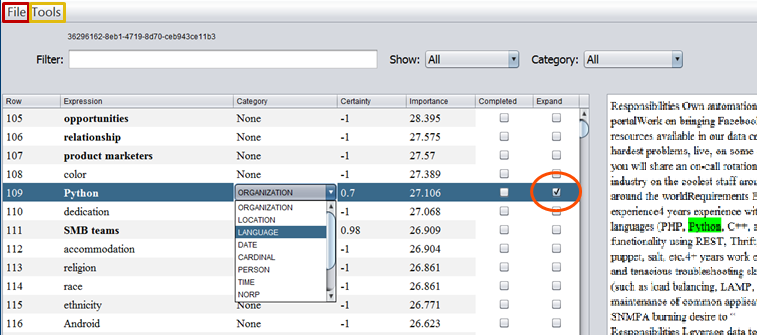}
\centering
\caption{\App user interface for seed selection and expansion.}
\label{app1_fig}
\end{figure*}

This section describes the iterative end-to-end workflow of \App, as depicted in Figure~\ref{flow_fig}. 
\paragraph{Steps 1 \& 2: Selecting an Input Corpus and Training Models.}
The first step of the flow is to \textbf{select an input corpus}, performed by selecting {\it Open} (not shown) from the {\it File} menu (see the red rectangle in Figure~\ref{app1_fig}). The second step of the flow is to \textbf{train the models} based on the selected corpus, performed by selecting {\it Train Models} (not shown) from the {\it Tools} menu (see the yellow rectangle in Figure~\ref{app1_fig}). The ``train models'' step extracts term groups from the corpus, trains the combined term groups embedding models (Section \ref{sec:TermRepresentation}) and the MLP classifier that predicts whether a candidate term should be part of the expanded set (Section \ref{sec:MultiContextSimilarity}).

\paragraph{Steps 3 \& 4: Selecting and Expanding a Seed Set.} 
Figure~\ref{app1_fig} also shows the seed set selection and expansion user interface. Each row in the displayed table corresponds to a different term group. The top 5000 term group names are displayed under the {\it Expression} column, sorted by their TF-IDF based importance score.
Term groups that include more than one term are highlighted in bold, and are represented in the display, by the term with the highest importance score among the terms of the group. Hovering over such a group opens a drop-down list that displays all the terms within the group. The user can choose to exclude specific terms from the group if their semantic meaning does not align with that of the group.
The {\it Filter} text box is used for searching for specific term groups.  
Upon selecting (clicking) a term group, the context view on the right hand side of Figure~\ref{app1_fig} (blurred) displays text snippets from the input corpus that include terms that are part of the selected term group (highlighted in green). 
The context view enables the user to verify the semantic meaning of terms in various contexts in the topical domain.

The user can \textbf{create a seed set} assembled from specific term groups by checking their {\it Expand} checkbox (see the red circle in Figure~\ref{app1_fig}). 
The user can set a name for the semantic category of the seed set. This name will be used for displaying and storing the seed set and the resulting expanded set of terms. The category name can be selected from a predefined list of category names or added as a new custom category name (see the drop down list in Figure~\ref{app1_fig}). 
Once the seed set is assembled, the user can \textbf{expand the seed set} by selecting the {\it Expand} option (not shown) in the {\it Tools} menu. 

\begin{figure}[ht]
\includegraphics[width=0.25\textwidth]{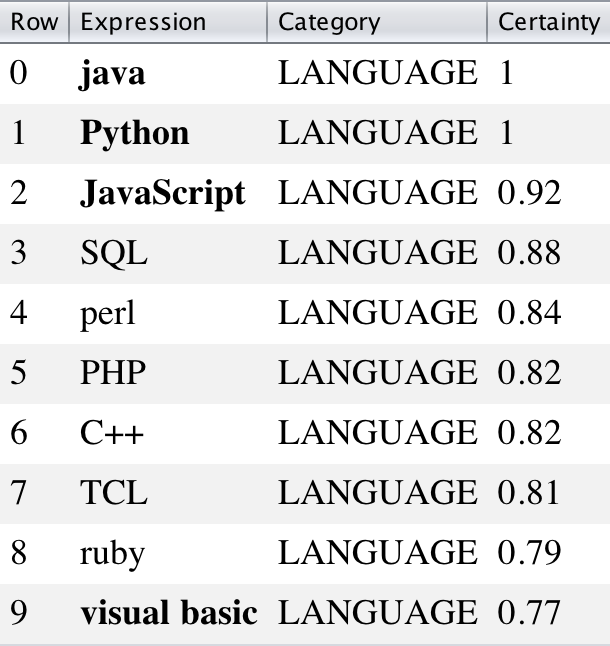}
\centering
\caption{\App user interface for expansion results output. Seed terms are `java' and `python'.
}
\label{app2_fig}
\end{figure}

\paragraph{Steps 5 \& 6: Edit, Validate and Re-expand.} 
Figure~\ref{app2_fig} shows the output of the expansion process. 
The {\it Certainty} score represents the relatedness of each expanded term group to the seed set, as determined by the MLP classifier (Section \ref{sec:MultiContextSimilarity}). The Certainty scores of term groups that were manually selected as part of the seed set are set to 1. 
The user can \textbf{validate} each expanded item by checking the {\it Completed} checkbox. The validated list can then be saved and later used as a fine-grained semantic class input to external applications. Following validation, the user can perform \textbf{re-expansion} by creating a new seed set based on the validated expanded terms and the original seed set terms. 

\section{Field Use Cases}\label{sec:usecases}
This section describes two use cases in which \App has been successfully used. 

\paragraph{Automated Recruitment System.} Human matching of applicant resumes to open positions in organizations is time-consuming and costly. Automated recruitment systems enable recruiters to speed up and refine this process. The recruiter provides an open position description and then the system scans the organization’s resume repository searching for the best matches. One of the main features that affect the matching is the skills list, for example, a good match between an applicant and an open position regarding specific programming skills or experience using specific tools is significant for the overall matching. However, manual generation and maintenance of comprehensive and updated skills lists is tedious and difficult to scale. \App was integrated into such a recruitment system. Recruiters used the system's user interface (Figures~\ref{app1_fig} \& \ref{app2_fig})  to generate fine-grained skills lists based on small seed sets for eighteen engineering job position categories.
We evaluated the recruitment system use case for different skill classes. The system achieved a precision of 94.5\%, 98.0\% and 70.5\% at the top 100 applicants, for the job position categories of Software Machine Learning Engineer, Firmware Engineer and ADAS Senior Software Engineer, respectively.

\paragraph{Issues and Defects Resolution.} Quick identification of duplicate defects is critical for efficient software development. The aim of automated issues and defects resolution systems is to find duplicates in large repositories of millions of software defects used by dozens of development teams. This task is challenging because the same defect may have different title names and different textual descriptions. The legacy solution relied on manually constructed lists of tens of thousands of terms, which were built over several weeks. Our term set expansion application was integrated into such a system and was used for generating domain specific semantic categories such as product names, process names, technical terms, etc. The integrated system enhanced the duplicate defects detection precision by more than 10\% and sped-up the term list generation process from several weeks to hours.

\section{Conclusion}
\label{sec:conclusion}
We presented \App, a corpus-based system for set expansion which enables users to select a seed set of terms, expand it, validate it, re-expand the validated set and store it. The expanded sets can then be used as a domain specific semantic classes for downstream applications. Our system was used in several real-world use cases, among them, an automated recruitment system and an issues and defects resolution system.

\bibliography{references.bib}
\bibliographystyle{acl_natbib_nourl}

\end{document}